\documentclass{article}
\usepackage{iclr2027_conference,times}
\usepackage[T1]{fontenc}
\usepackage{graphicx}
\usepackage{amsmath,amssymb}
\usepackage{booktabs,array,multirow,makecell,colortbl}
\usepackage{caption}
\usepackage{algorithm,algorithmic,newfloat,listings}
\usepackage{hyperref}
\usepackage{xurl}
\usepackage{placeins}
\usepackage{wrapfig}
\usepackage{needspace}
\DeclareCaptionStyle{ruled}{labelfont=normalfont,labelsep=colon,strut=off}
\floatstyle{ruled}
\newfloat{listing}{tb}{lst}{}
\floatname{listing}{Listing}

\newcommand{\ourframework}{MemScope}
\newcommand{\flashicon}{\raisebox{-0.13em}{\includegraphics[height=0.95em]{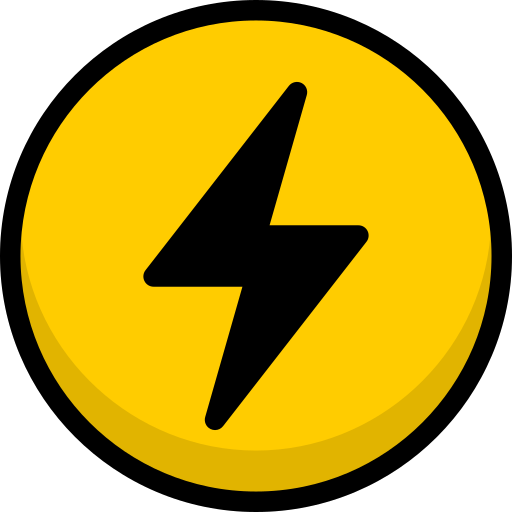}}}

\newcommand{\etc}{etc.}

\makeatletter
\def\input@path{{aaai_original/}}
\makeatother
\hypersetup{pdftitle={MemScope: Evolving Agent Memory Architectures through Localized Code Repair},pdfauthor={Miao Su, Yucan Guo, Zixuan Li, Long Bai, Zhongni Hou, Xiaolong Jin, Jiafeng Guo},colorlinks=true,linkcolor=blue,citecolor=blue,urlcolor=blue}

\title{MemScope: Evolving Agent\\Memory Architectures through\\Localized Code Repair}
\author{Miao Su$^{1,2,3}$, Yucan Guo$^{1,2,3}$, Zixuan Li$^{1,2,3}$, Long Bai$^{1,2,3}$ \\
\textbf{Zhongni Hou$^{4}$, Xiaolong Jin$^{1,2,3}$, Jiafeng Guo$^{1,2,3}$} \\
$^{1}$State Key Laboratory of AI Safety \\
$^{2}$Institute of Computing Technology, Chinese Academy of Sciences \\
$^{3}$University of Chinese Academy of Sciences \\
$^{4}$Meituan}
\iclrfinalcopy
\begin{document}
\maketitle
\lhead{Preprint}
\begin{abstract}
Agent memory architectures define how LLM agents store and reuse experience across tasks, but they are typically fixed after deployment, limiting their adaptability across task types. Recent self-evolving approaches instead adapt these memory architectures using execution feedback, typically represented as evaluation scores or compressed summaries. However, such coarse-grained feedback is insufficient for precise failure localization in current agent systems, where long-horizon task execution involves multiple interacting components. Without a precise failure location, existing approaches lack a clear code-level target for modification, leaving the update scope weakly constrained and often leading to large-scale rewrites of an otherwise capable memory architecture. We propose \ourframework{}, a framework for evidence-grounded, localized evolution of an existing memory architecture. Its tool-using Patch Agent jointly inspects raw execution traces, runtime memory state, and current architecture code to identify repair targets and apply bounded, in-place edits under provider-identity and memory-compatibility constraints. On GAIA, the trace-and-memory inspection variant achieves 86.67\% accuracy, compared with 70.00\% for trace-only and 63.33\% for summary-only feedback. Under xBench tasks, the evolved memory architecture improves over its original version from 49 to 55 correct answers with DeepSeek V3.2 and from 50 to 53 with Kimi K2.7 Code. On disjoint ALFWorld unseen tasks, success rises from 50.75\% to 71.64\%. Across archived architecture-snapshot transitions, the median code-change ratio is 4.4\%. These results support fine-grained execution evidence as a basis for localized, in-place memory-architecture evolution.
\end{abstract}


\section{Introduction}
\label{sec:introduction}

An LLM agent combines a language model with an agent harness that manages tool use, context construction, execution control, and memory~\citep{weng2026harness,lee2026metaharness,zhang2026selfharness,lin2026ahe}. Within this harness, the memory architecture implements reusable decisions about what experience to store, how to retrieve it, and how to present it to the agent~\citep{zhongMemoryBankEnhancingLarge2023,chhikaraMem0BuildingProductionReady2025,wangAgentWorkflowMemory2024,duRethinkingMemoryAI2025a}. Most existing architectures remain fixed after deployment even as agents encounter open-ended tasks with different memory demands. This motivates adapting not only the contents of memory, but also the executable mechanisms that manage them.

Recent approaches use execution outcomes to evolve agent memory, but differ in what they allow to change. Some update the experience retained by the agent while leaving the executable logic for storage, retrieval, and presentation fixed~\citep{shinn2023reflexion,liu2025cer,fangMempExploringAgent2025,ouyangReasoningBankScalingAgent2025,zhang2026memskilllearningevolvingmemory,zhangMemRLSelfEvolvingAgents2026}. Such content-level updates cannot directly correct failures caused by the logic itself. Architecture-level methods instead make the executable memory architecture mutable, allowing evolution to modify the code that determines what is stored, how it is retrieved, and how retrieved information is presented~\citep{zhang2025memevolvemetaevolutionagentmemory,xiong2026learningcontinuallylearn}. This shifts the evolution target from memory contents to the mechanisms that operate on them.

\begin{figure*}[!t]
  \centering
  \includegraphics[width=\textwidth]{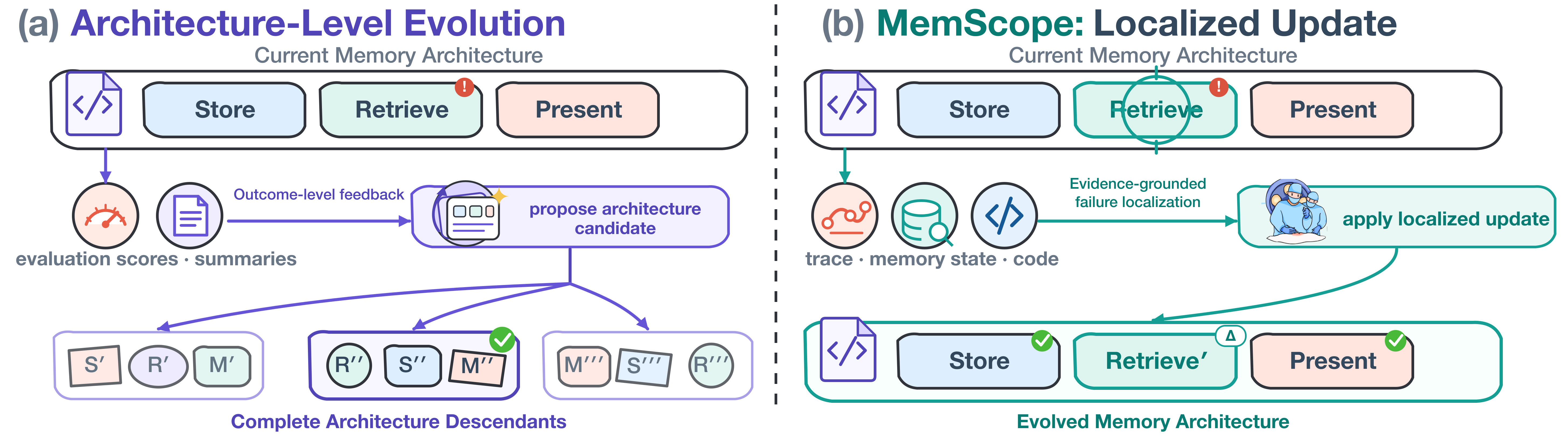}
  \caption{\textbf{Feedback and update granularity in memory-architecture evolution.} Prior architecture-level approaches use outcome-level feedback to propose complete executable architectures. \ourframework{} makes the repair target explicit by jointly inspecting execution traces, runtime memory state, and provider code, then applies a localized, in-place edit to the existing provider. In the illustrated update, only the highlighted code region is modified.}
  \label{fig:introduction}
\end{figure*}

We focus on a complementary problem: identifying where an existing memory architecture should change. Evaluation scores and compressed summaries do not directly connect an observed failure to a specific implementation region. In long-horizon execution, this connection is difficult because behavior depends on interactions between stored experience, retrieval decisions, and downstream agent actions. Figure~\ref{fig:introduction} contrasts complete-architecture proposals driven by outcome-level feedback~\citep{xiong2026learningcontinuallylearn,zhang2025memevolvemetaevolutionagentmemory} with evidence-grounded identification of a local repair target.

Consider an agent that fails to receive useful memory guidance. The relevant experience may be absent from memory, excluded by retrieval filtering, or bypassed by an early-return code path. These explanations suggest different interventions; a failure may also arise after suitable guidance has already been delivered and require no provider change. Inspecting raw traces, runtime memory state, and current code can help distinguish these cases. We therefore investigate whether fine-grained execution evidence can support targeted, in-place updates to an existing memory architecture.

To study this question, we introduce \ourframework{}, a framework for evidence-grounded, localized evolution of agent memory architectures. An Agent Executor performs tasks with the active architecture, while a tool-using Patch Agent inspects execution evidence between task segments. Each architecture is packaged as a reusable memory skill with a natural-language \texttt{SKILL.md} file and an executable provider, separate from its runtime memory bank~\citep{anthropic_skills_2025}. When repair is triggered, the Patch Agent examines successful and failed raw traces, stored memories and retrieval decisions, and current provider code to identify a candidate repair region. It then proposes exact old-to-new replacements under a local-update contract that preserves provider identity and requires compatibility with existing memory records. A patch that passes local checks is installed for subsequent execution, retaining the runtime memory bank; insufficient evidence can instead lead to a no-patch decision.

Under DeepSeek V3.2, MemScope + Flash-Searcher reaches 83.64\% accuracy on GAIA validation and 78\% on complete xBench, compared with 67.88\% and 70\% for MemEvolve + Flash-Searcher; MemScope's complete xBench score with Kimi K2.7 Code is 74\%, and a separate matched-start comparison favors the evolved DynamicCheatsheet over the original with both DeepSeek (55/70 versus 49/70) and Kimi (53/70 versus 50/70). On ALFWorld, unseen-task success after the same online-memory collection is 71.64\% for MemScope versus 50.75\% for the original DynamicCheatsheet, while the GAIA diagnostic ablation yields 86.67\% for trace-and-memory inspection versus 63.33\% for summary-only feedback. Across 14 archived provider transitions, the median code-change ratio is 4.4\%.

Our contributions are:
\begin{itemize}
  \item We introduce an evidence-grounded repair interface that jointly exposes raw execution traces, runtime memory state, and provider code, making failure localization an explicit step in the evolution of an existing memory architecture.
  \item We develop \ourframework{} to turn the resulting diagnosis into bounded, in-place code edits under provider-identity and memory-compatibility constraints. Across 14 archived provider-snapshot transitions, the median code-change ratio is 4.4\%.
  \item In the GAIA diagnostic ablation, trace-and-memory inspection yields 86.67\% accuracy versus 63.33\% for summary-only feedback. Matched-start xBench comparisons favor the evolved provider under both backbones, and ALFWorld unseen-task success improves from 50.75\% to 71.64\% after a common memory-collection sequence.
\end{itemize}

\section{Related Work}
\label{sec:related-work}

\subsection{Agent Memory}
\label{sec:rw-agent-memory}

Agent memory supports personalization and experience reuse~\citep{duRethinkingMemoryAI2025a,zhangSurveyMemoryMechanism2024,packerMemGPTLLMsOperating2024,rasmussenZepTemporalKnowledge2025a}. Its development spans trace storage, reflection, and experience abstraction~\citep{luo2026storageexperience}, with memory behavior evaluated within and across episodes by EvoMemBench~\citep{wang2026evomembench}. MemoryBank~\citep{zhongMemoryBankEnhancingLarge2023}, Mem0~\citep{chhikaraMem0BuildingProductionReady2025}, and temporal semantic memory~\citep{suDialogueTimeTemporal2026} retain preferences, temporal states, and evolving personas for long-term coherence. Experience-based systems instead reuse execution traces: Agent Workflow Memory induces web-navigation workflows~\citep{wangAgentWorkflowMemory2024}, while Memp distills procedural memories~\citep{fangMempExploringAgent2025,huHiAgentHierarchicalWorking2024}. Most retain a fixed memory architecture after deployment.

Yet retrieved traces can propagate errors or encourage misaligned replay~\citep{xiongHowMemoryManagement2025,chenHaluMemEvaluatingHallucinations2025}. MemScope adapts the architecture itself: provider code defines task-family storage and retrieval mechanisms, while a skill document specifies their use for the current task stream.

\subsection{Self-Evolving Agents}
\label{sec:rw-self-evolving-agents}

Self-evolving agents use execution feedback to improve persistent components, including solution-generating programs, agent implementations, and execution harnesses~\citep{zelikman2024stop,zhang2025dgm,lee2026metaharness,zhang2026selfharness,lin2026ahe}. Many memory-related approaches accumulate or refine non-parametric experience and natural-language memory skills~\citep{shinn2023reflexion,liu2025cer,fangMempExploringAgent2025,zhang2026memskilllearningevolvingmemory,ouyangReasoningBankScalingAgent2025,zhangMemRLSelfEvolvingAgents2026,zhangMemGenWeavingGenerative2025a,linMUSEAutoskillSelfEvolvingAgents2026,ouyangSkillOSLearningSkill2026a,yangAutoSkillExperienceDrivenLifelong2026,liSkillHoneHarnessContinual2026}, changing retained content or memory-construction instructions while keeping the executable architecture fixed. Evo-Memory studies memory revision over sequential task streams~\citep{wei2025evomemory}, while FORGE turns failure traces into textual memories shared among agents~\citep{bogdanov2026forge}. TMEM combines explicit memory with online parametric adaptation~\citep{ren2026parametricmemory}; AutoAgent evolves cognition and orchestrates elastic memory~\citep{wang2026elasticmemory}.

ALMA searches complete memory architectures expressed as code, and MemEvolve jointly evolves experiential knowledge and modular memory architectures~\citep{xiong2026learningcontinuallylearn,zhang2025memevolvemetaevolutionagentmemory}. MemScope's local-evolution path takes an existing provider as its starting point and uses trace- and memory-grounded replacements, making the repair target and edit scope explicit.

LLM-based automated program repair studies how models diagnose software failures and generate code changes~\citep{yang2024sweagent}, with repository-level repair evaluated on benchmarks such as SWE-bench~\citep{jimenez2024swebench}. MemScope applies localized repair to the agent's own memory architecture, grounding updates in task execution traces, runtime memory, and provider code.

\begin{figure*}[!t]
  \centering
  \includegraphics[width=\textwidth]{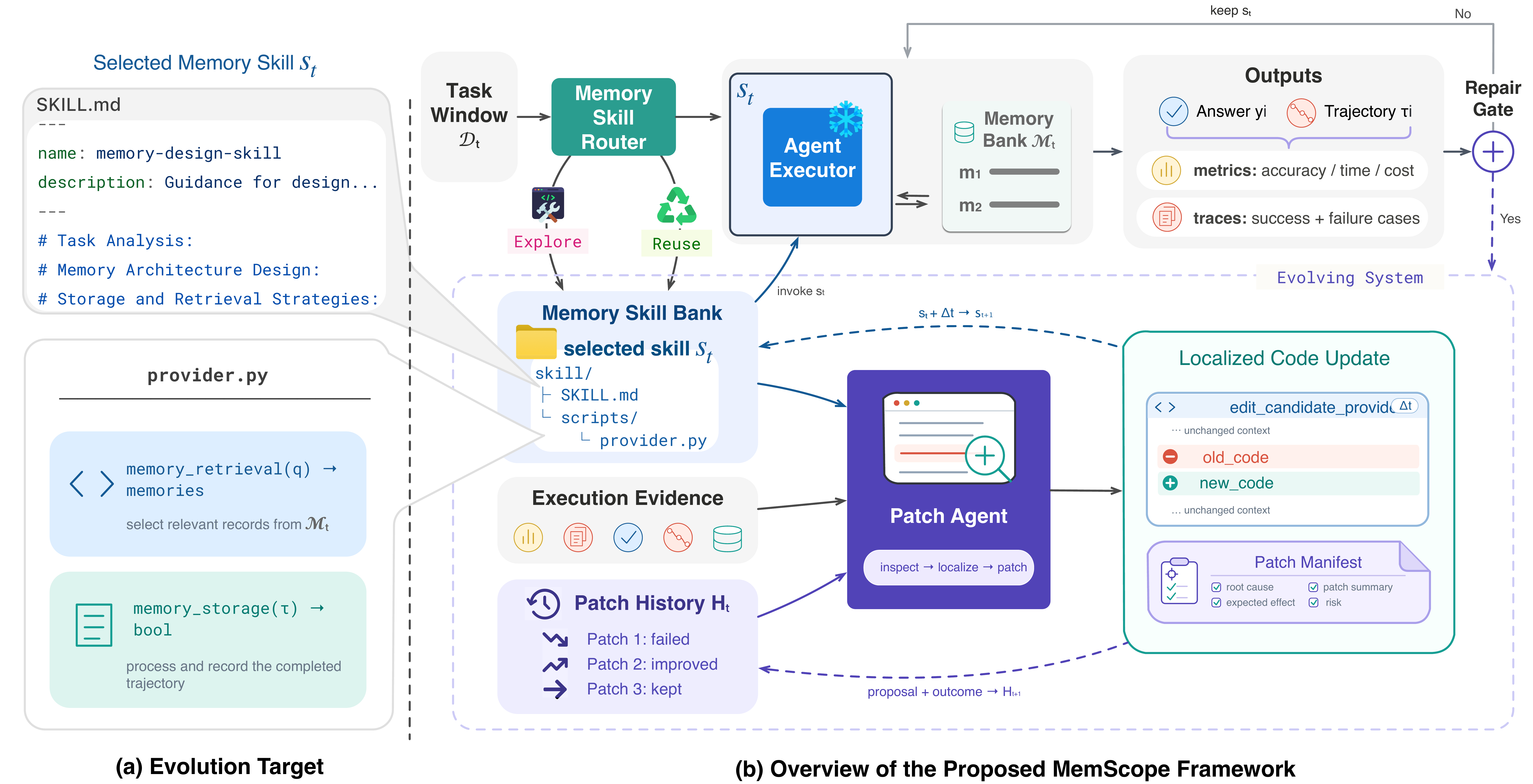}
  \caption{Overview of \ourframework{}. (a) Evolution Target. Following the Agent Skills format, each reusable memory skill contains a required \texttt{SKILL.md} file and an executable provider under \texttt{scripts/}; the activated provider maintains a separate skill-specific runtime memory bank $\mathcal{M}_t$. (b) MemScope Evolution Loop. Execution produces metrics and raw success and failure traces. When repair is triggered, the Patch Agent inspects these traces, the active memory bank, the current provider, and Patch History. A local edit changes $p_t$ into $p_{t+1}$ without replacing the skill, while the patch and its later outcome are recorded for subsequent evolution.}
  \label{fig:method}
\end{figure*}


\section{Method}
\label{sec:method}

Figure~\ref{fig:method} shows the evolution target and loop. For an existing skill, MemScope uses task evidence to repair the active provider in place when the trigger fires. Provider code changes locally while the separate runtime memory continues to update during task use. Our provider-evolution runs start from the existing DynamicCheatsheet memory architecture~\citep{suzgun2025_DynamicCheatsheet} and distinguish diagnostic-evidence variants from original-versus-evolved provider comparisons.

\subsection{Evolution Target}
\label{sec:method-target}

Each memory skill $s_t$ combines natural-language guidance in \texttt{SKILL.md} with an executable provider $p_t$ implementing \texttt{memory\_retrieval} and \texttt{memory\_storage}, following the Agent Skills format~\citep{anthropic_skills_2025}. MemScope locally updates $p_t$ and records the design lesson in \texttt{SKILL.md}; runtime memory remains separate from the reusable skill (Figure~\ref{fig:method}(a)). A recorded diagnosis and the Patch-Agent prompt appear in Appendices~\ref{app:compact-details} and~\ref{app:prompts}.

\subsection{MemScope Evolution Loop}
\label{sec:method-search-loop}

As shown in Figure~\ref{fig:method}(b), an evolution episode is one intended pass over a task window $\mathcal{D}_t$. MemScope first obtains probe evidence, uses a fixed or routed skill, and partitions the remaining tasks into sequential segments executed with the active provider. After each segment, the repair trigger determines whether to invoke the Patch Agent.

\noindent\textbf{Skill selection and task execution.} MemScope initializes the Memory Skill Bank with 12 human-designed memory skills from the EvolveLab codebase of MemEvolve~\citep{zhang2025memevolvemetaevolutionagentmemory}. At the start of each episode, MemScope runs probe tasks from $\mathcal{D}_t$. The Memory Skill Router uses the probe tasks and their execution evidence together with the available skill descriptions to select a suitable existing skill $s_t$ and activate its provider $p_t$. When the Router determines that no existing skill is suitable, MemScope enters global exploration: the Skill Designer derives a new skill specification from the probe evidence, and MemScope generates a complete provider from this specification before registering and evaluating the new skill-provider pair.

After routing, the frozen Agent Executor $\mathcal{A}$ executes the remaining tasks using the selected provider $p_t$ and a skill-specific runtime memory bank $\mathcal{M}_t$, which stores task experience separately from the reusable skill files. For each task $x_i$, the provider calls \texttt{memory\_retrieval} to obtain relevant memories, and the Agent Executor produces an answer $y_i$ and trajectory $\tau_i$. MemScope retains $\tau_i$ as a raw execution trace for diagnosis, while the provider applies \texttt{memory\_storage} to distill reusable information from $\tau_i$ into a memory item $m_i$ and update the runtime memory bank:
\begin{align}
  r_i &= \operatorname{memory\_retrieval}_{p_t}(x_i,\mathcal{M}_t), \\
  (y_i,\tau_i) &= \mathcal{A}(x_i,r_i), \\
  m_i &= \operatorname{memory\_storage}_{p_t}(\tau_i), \\
  \mathcal{M}_t &\leftarrow \mathcal{M}_t \oplus m_i.
  \label{eq:memory-execution}
\end{align}
Here, $r_i$ denotes retrieved memories, $m_i$ the distilled memory item, and $\oplus$ provider-specific insertion or update. The runtime bank is retained across a localized patch. Sequential execution segments yield accuracy, time, cost, and raw success/failure traces for the repair trigger and Patch Agent.

\noindent\textbf{Repair trigger.} MemScope tracks the best execution accuracy from previous episodes on the same task window. Let $a_t$ be current segment accuracy, $a_t^\star$ this historical reference, and $f_t$ the current number of failures. When performance falls below the reference by more than $\delta$ and at least $f_{\min}$ tasks fail, MemScope invokes the Patch Agent to inspect the execution evidence: $g_t=\mathbb{I}[a_t^\star-a_t>\delta]\,\mathbb{I}[f_t\geq f_{\min}]$. We use $\delta=0.05$ and $f_{\min}=1$. Patches that pass local checks update the provider for subsequent execution; failed checks and \texttt{no\_patch} retain $p_t$. Table~\ref{tab:no-patch-case} illustrates a recorded \texttt{no\_patch} decision.

\FloatBarrier
\subsection{Patch Agent: Fine-Grained Diagnosis and Localized Updates}
\label{sec:method-local-update}

The Patch Agent makes failure localization an explicit step in memory-architecture evolution, turning observed task failures into evidence-grounded, localized updates to the active provider. As shown in Figure~\ref{fig:method}(b), it takes execution evidence, the selected memory skill \(s_t\), and Patch History \(\mathcal{H}_t\) as inputs, and produces a localized code update \(\Delta_t\) together with a patch manifest that summarizes the proposed root cause, patch, expected effect, and risk.

\begin{table}[t]
\caption{Evidence sources for failure localization.}
\label{tab:diagnostic-tools}

\centering
\small
\setlength{\tabcolsep}{3pt}
\renewcommand{\arraystretch}{1.0}
\begin{tabular}{@{}>{\raggedright\arraybackslash}p{0.46\columnwidth}>{\raggedright\arraybackslash}p{0.48\columnwidth}@{}}
\toprule
\textbf{Tool or action} & \textbf{Evidence exposed} \\
\midrule
\multicolumn{2}{@{}l}{\textit{Trace Inspection}} \\
\path{list_traces}; \path{read_trace_header}; \path{read_trace_steps}; \path{grep_traces} & Index available cases; inspect a task result and step map; read selected native steps; search matching events across traces. \\
\addlinespace[2pt]
\multicolumn{2}{@{}l}{\textit{Memory Inspection (\texttt{memory\_store})}} \\
\texttt{stats} & Summarize record counts, fields, schemas, and outcome distribution. \\
\texttt{list} & List compact record summaries, optionally filtered by outcome. \\
\texttt{read} & Return one complete raw memory record. \\
\texttt{query} & Execute a recorded request and return raw matches, selection decisions, and the retrieval result. \\
\addlinespace[2pt]
\multicolumn{2}{@{}l}{\textit{Code Grounding}} \\
\path{read_candidate_provider}; \path{read_skill_file} & Inspect the current executable code and the skill's intended scope and memory-architecture design. \\
\bottomrule
\end{tabular}

\end{table}

\noindent\textbf{Diagnosis and localized update.} The Patch Agent inspects execution traces, runtime memory, and provider code through the read-only tools in Table~\ref{tab:diagnostic-tools} to identify a repair target. A score alone cannot distinguish absent experience, filtering/ranking errors, and ineffective guidance formatting; the tools expose observed behavior, stored experience, retrieval decisions, and implementation for on-demand inspection. The GAIA ablation in Section~\ref{sec:evolution-ablation} compares summary-only feedback with access to raw traces and memory inspection, using task accuracy as the outcome. The Patch Agent may use the tools repeatedly and in any order to select a proposed edit location. Successful traces provide contrastive evidence for selecting the repair target.

Having selected a repair target, the Patch Agent emits exact old-to-new replacements on an experiment-local provider and a manifest containing its hypothesized root cause, expected effect, and risk. Early GAIA artifacts check syntax, class identity, memory type, and required method names; later xBench/ALFWorld artifacts additionally record import and isolated-retrieval checks. Prompts require schema compatibility and prohibit broad rewrites and task/answer hardcoding. After local validation, the updated provider is used in subsequent task execution, and its observed outcomes are recorded in Patch History.

\noindent\textbf{Patch history.} Patch History $\mathcal{H}_t$ records each proposal, its installation decision, and later outcomes on revisited tasks, including recovered failures and regressed successes. The Patch Agent can consult these records when making subsequent repair decisions on the same provider and skill. The xBench and ALFWorld cases in Section~\ref{sec:evolution-ablation} illustrate this use.
\FloatBarrier


\section{Experiments}
\label{sec:experiments}

\subsection{Experimental Setup}
\label{sec:experimental-protocol}

\paragraph{Benchmarks.}
Following prior evaluations of general-purpose and self-evolving agents~\citep{laws2024trase,qiu2025alita,zhang2025memevolvemetaevolutionagentmemory}, we evaluate MemScope on three benchmarks. GAIA~\citep{mialon2023gaia} covers real-world tasks requiring multi-step reasoning, web browsing, multimodal understanding, and tool use; its validation set contains 165 tasks across three difficulty levels. xBench DeepSearch-2505~\citep{xbench2025} evaluates web research and information synthesis. ALFWorld~\citep{shridhar2021alfworld} evaluates sequential household interaction in a text-based environment. Together, they cover information-seeking and embodied task settings. Dataset-specific evaluation protocols appear below.

\paragraph{Evolution and evaluation.}
We use Flash-Searcher~\citep{qin2025flash} as the Agent Executor for GAIA and xBench, with DeepSeek V3.2~\citep{deepseekai2025deepseekv32} and Kimi K2.7 Code~\citep{kimiteam2025kimik2} as backbone models. For GAIA tasks with visual inputs, Flash-Searcher uses GPT-4o~\citep{openai2024gpt4o} (\texttt{gpt-4o-2024-11-20}) for visual inspection. ALFWorld uses DeepSeek V3.2 with a text-action executor. Providers use \texttt{all-MiniLM-L6-v2}~\citep{reimers2019sentencebert,wang2020minilm} for semantic retrieval. GAIA and xBench answers are evaluated by GPT-4o-mini~\citep{openai2024gpt4omini}; ALFWorld uses environment-reported task success.

Across the three benchmarks, the intended evolution schedule comprises five episodes over 30 development tasks, with three probes and 27 execution tasks in segments of 10, 10, and 7. Following prior work~\citep{zhang2025memevolvemetaevolutionagentmemory,xiong2026learningcontinuallylearn}, we use sequential online evaluation to measure the resulting memory systems during continued use. Following MemEvolve~\citep{zhang2025memevolvemetaevolutionagentmemory}, we evaluate GAIA on the full validation set. For xBench and ALFWorld, code persists across evolution episodes while memory resets between episodes and accumulates within each; post-evolution comparisons use the last code state and continue updating memory. The separate xBench fixed-code comparison starts original and evolved providers from independent empty stores, as does ALFWorld before its common collection sequence. In these comparisons, the Agent Executor receives memory retrieved by the provider without injecting \texttt{SKILL.md} experience into its task context. Our xBench and ALFWorld outcomes count raw \texttt{score==1} records, with errors retained in the denominator.

\subsection{Main Results on GAIA and xBench}
\label{sec:self-evolving-comparison}
\label{sec:xbench}

\begin{table*}[!t]
\caption{GAIA validation and xBench DeepSearch-2505 accuracy (\%).}
\label{tab:main-table-1}
\label{tab:xbench-main}
\centering
\small
\setlength{\tabcolsep}{3pt}
\renewcommand{\arraystretch}{1.0}
\begin{tabular*}{\textwidth}{@{\extracolsep{\fill}}>{\raggedright\arraybackslash}p{0.35\textwidth}>{\raggedright\arraybackslash}p{0.16\textwidth}|ccccc@{}}
\toprule
\multirow{2}{*}{\textbf{System}} & \multirow{2}{*}{\textbf{Model Family}} & \multicolumn{4}{c}{\textbf{GAIA}} & \multirow{2}{*}{\textbf{xBench}} \\
\cmidrule(lr){3-6}
& & \textbf{Avg.} & \textbf{L1} & \textbf{L2} & \textbf{L3} & \\
\midrule
\multicolumn{7}{l}{\textit{Closed-Source Agent Frameworks}} \\
Alita (pass@1) & Claude Sonnet 4, GPT-4o & 75.15 & 77.36 & 76.74 & 65.38 & --- \\
Langfun Agent & Claude 3.7 \etc{} & 71.52  & 83.02 &  68.60 & 57.69 & --- \\
TraseAgent & Claude \etc{} & 70.30 & 83.02 & 69.77 & 46.15 & --- \\
OpenAI Deep Research & o1, o3 \etc{} & 67.36 & 74.29 & 69.06 & 47.60 & --- \\
h2oGPTe & Claude 3.5 & 63.64 & 67.92 & 67.44 & 42.31 & --- \\
Desearch & GPT-4o & 56.97 & 71.70 & 58.14 & 23.08 & --- \\
OMNE & o1-preview & 46.06 & 60.38 & 44.19 & 23.08 & --- \\
\midrule
\multicolumn{7}{l}{\textit{Open-Source Agent Frameworks}} \\
OWL Workforce (pass@3) & GPT-4o+o3-mini & 60.61 & 81.14 & 58.14 & 26.92 & 55.00 \\
OWL Roleplaying (pass@3) & GPT-4o+o3-mini & 58.18 & 81.14 & 54.65 & 23.08 & --- \\
TapeAgents & Claude 3.7 \etc{} & 55.76 & 71.70 & 53.49 & 30.77 & --- \\
AutoAgent & Claude 3.5 \etc{} & 55.15 & 71.70 & 53.40 & 26.92 & --- \\
Magentic-One & OpenAI o1 \etc{} & 46.06 & 56.60 & 46.51 & 23.08 & --- \\
Cognitive Kernel-Pro (pass@1) & Claude 3.7 \etc{} & 60.00 & 79.25 & 56.98 & 30.77 & 56.00 \\
Cognitive Kernel-Pro (pass@3) & Claude 3.7 \etc{} & 75.15 & 84.91 & 73.26 & 61.54 & --- \\
OAgents & Claude 3.7 \etc{} & 66.67 & 77.36 & 66.28 & 46.15 & 47.00 \\
Agent KB (pass@1) & GPT-4.1 & 61.21 & 79.25 & 58.14 & 34.62 & 48.00 \\
Agent KB (pass@2) & GPT-4.1 & 67.27 & 83.02 & 67.44 & 34.62 & 58.00 \\
Agent KB (pass@3) & GPT-4.1 & 73.94 & 84.91 & 73.26 & 53.85 & 68.00 \\
Flash-Searcher \flashicon{} (pass@1) & Kimi K2 & 52.12 & 58.49 & 52.33 & 34.62 & 66.00 \\
Flash-Searcher \flashicon{} (pass@1) & DeepSeek V3.2 & 60.61 & 79.25 & 53.49 & 46.15 & 68.00 \\
\midrule
\multicolumn{7}{l}{\textit{Self-Evolving Memory Architectures}} \\
\addlinespace[1.5pt]
MemEvolve + \flashicon{} (pass@1) & Kimi K2 & 61.21 & 67.92 & 63.95 & 38.46 & 68.00 \\
MemEvolve + \flashicon{} (pass@1) & DeepSeek V3.2 & 67.88 & 83.02 & 63.95 & 50.00 & 70.00 \\
\midrule
\textbf{\ourframework{} + \flashicon{}} (pass@1) & Kimi K2.7 Code & 78.18 & 88.68 & 75.58 & 65.38 & 74.00 \\
\textbf{\ourframework{} + \flashicon{}} (pass@1) & DeepSeek V3.2 & \textbf{83.64} & \textbf{90.57} & \textbf{81.40} & \textbf{76.92} & \textbf{78.00} \\
\bottomrule
\end{tabular*}
\end{table*}

\noindent\textbf{Closed-source agent frameworks.} On GAIA, \ourframework{} + Flash-Searcher with DeepSeek V3.2 reaches 83.64\% pass@1 accuracy, compared with 75.15\% pass@1 for Alita, the highest result in the closed-source group (Table~\ref{tab:main-table-1}).

\noindent\textbf{Open-source agent frameworks.} Cognitive Kernel-Pro has the highest GAIA average in this group at 75.15\% pass@3; \ourframework{} reaches 83.64\% pass@1. Under DeepSeek V3.2 and pass@1, \ourframework{} + Flash-Searcher improves on Flash-Searcher from 60.61\% to 83.64\% on GAIA and from 68\% to 78\% on xBench.

\noindent\textbf{Self-evolving memory architectures.} Under DeepSeek V3.2, MemEvolve + Flash-Searcher reaches 67.88\% on GAIA and 70\% on xBench, while \ourframework{} + Flash-Searcher reaches 83.64\% and 78\%, respectively. The ordering is consistent across both benchmarks.

With Kimi K2.7 Code, \ourframework{} reaches 78.18\% on GAIA and 74\% on xBench (Table~\ref{tab:main-table-1}).

Following MemEvolve~\citep{zhang2025memevolvemetaevolutionagentmemory}, we evaluate xBench on the full DeepSearch-2505 set using a 40-step budget and GPT-4o-mini judging. To show how often evolution retains a prior success, recovers a failure, or causes a regression, Table~\ref{tab:xbench-paired-outcomes} pairs original and final evolved DynamicCheatsheet outcomes on the same 70 questions (31--100). Within each backbone, both providers use Flash-Searcher and start with independent empty memory stores that update online from trajectories and post-task correctness feedback, without the judge's reference answer. The evolved versus original providers answer 55/70 versus 49/70 correctly with DeepSeek V3.2 and 53/70 versus 50/70 with Kimi K2.7 Code.

\begin{table}[!htbp]
\centering
\caption{Paired xBench outcomes on 70 questions. C/I denote correct/incorrect; transitions are from the original to the evolved provider.}
\label{tab:xbench-paired-outcomes}
\small
\setlength{\tabcolsep}{6pt}
\renewcommand{\arraystretch}{1.1}
\begin{tabular*}{\linewidth}{@{\extracolsep{\fill}}lcccc@{}}
\toprule
\textbf{Backbone} & \textbf{C $\rightarrow$ C} & \textbf{I $\rightarrow$ C} & \textbf{C $\rightarrow$ I} & \textbf{I $\rightarrow$ I} \\
\midrule
DeepSeek V3.2 & 47 & 8 & 2 & 13 \\
Kimi K2.7 Code & 46 & 7 & 4 & 13 \\
\bottomrule
\end{tabular*}
\end{table}

Table~\ref{tab:xbench-paired-outcomes} decomposes the gains into retained successes, recovered failures, and regressions. The evolved provider retains 95.92\% and 92.00\% of original successes under DeepSeek and Kimi, respectively, with recovered failures outnumbering regressions in both runs.

\subsection{Generalization to Unseen ALFWorld Tasks}
\label{sec:alfworld}

\begin{wraptable}{r}{0.43\textwidth}
\centering
\caption{ALFWorld \texttt{valid\_unseen} success under DeepSeek V3.2 (correct/67, \%).}
\label{tab:alfworld-main}
\small
\setlength{\tabcolsep}{2pt}
\renewcommand{\arraystretch}{1.0}
\begin{tabular*}{\linewidth}{@{\extracolsep{\fill}}lc@{}}
\toprule
\textbf{System} & \textbf{Success} \\
\midrule
DynamicCheatsheet & 34/67 (50.75) \\
ALMA & 45/67 (67.16) \\
\midrule
\textbf{\ourframework{}} & \textbf{48/67 (71.64)} \\
\bottomrule
\end{tabular*}
\end{wraptable}

Following ALMA~\citep{xiong2026learningcontinuallylearn}, we evaluate ALFWorld on the full \texttt{valid\_unseen} set, disjoint from the evolution tasks. All methods start with empty memory and follow the same memory-collection sequence before evaluation, with memory updating online. The comparison uses DeepSeek V3.2, temperature zero, 32-token action outputs, 30 outer agent steps, and a 50-step environment cap. Original and evolved DynamicCheatsheet share the runner and task manifests; the ALMA row evaluates an existing design under this task sequence.

Under this shared local evaluation, \ourframework{} succeeds on 48/67 ALFWorld unseen tasks (71.64\%), versus 34/67 (50.75\%) for the original DynamicCheatsheet~\citep{suzgun2025_DynamicCheatsheet} and 45/67 (67.16\%) for ALMA~\citep{xiong2026learningcontinuallylearn} (Table~\ref{tab:alfworld-main}). The same original-versus-evolved ordering appears in xBench's matched-start comparison, extending the evidence from web research to unseen sequential-action tasks.

\subsection{Ablation and Analysis}
\label{sec:evolution-ablation}

\begin{table*}[!t]
\caption{GAIA diagnostic ablation: accuracy (\%), mean time, and steps.}
\label{tab:diagnostic-ablation}

\centering
\scriptsize
\providecommand{\ablateyes}{\ensuremath{\checkmark}}
\providecommand{\ablateno}{\ensuremath{\times}}
\setlength{\tabcolsep}{2.7pt}
\renewcommand{\arraystretch}{1.03}
\begin{tabular*}{\textwidth}{@{\extracolsep{\fill}}p{0.15\textwidth}ccccrrrrrr@{}}
\toprule
\multicolumn{2}{c}{\textbf{Experimental setting}} &
\multicolumn{3}{c}{\textbf{Diagnostic inputs}} &
\multicolumn{4}{c}{\textbf{Task accuracy (\%)}} &
\multicolumn{2}{c}{\textbf{Resources}} \\
\cmidrule(lr){1-2}\cmidrule(lr){3-5}\cmidrule(lr){6-9}\cmidrule(lr){10-11}
\textbf{Setting} & \textbf{Provider} & \textbf{Evidence} & \textbf{Memory} &
\textbf{History} & \textbf{Overall} & \textbf{L1} &
\textbf{L2} & \textbf{L3} & \textbf{Time (s)} & \textbf{Steps} \\
\midrule
No Memory & \ablateno & None & \ablateno & \ablateno & 56.67 & 75.0 & 58.82 & 20.0 & 517.61 & 11.30 \\
Original Provider & \ablateyes & None & \ablateno & \ablateno & 56.67 & 62.5 & 64.71 & 20.0 & 771.05 & 32.53 \\
\midrule
Summary-only & \ablateyes & Summary & \ablateno & \ablateno & 63.33 & 75.0 & 70.59 & 20.0 & 993.14 & 40.17 \\
Trace-only & \ablateyes & Raw Trace & \ablateno & \ablateno & 70.00 & \textbf{87.5} & 70.59 & 40.0 & 921.93 & 30.37 \\
\makecell[l]{Trace + Memory\\Inspection} & \ablateyes & Raw Trace & \ablateyes & \ablateno & \textbf{86.67} & 75.0 & \textbf{94.12} & \textbf{80.0} & 804.41 & 25.73 \\
\midrule
\ourframework{} & \ablateyes & Raw Trace & \ablateyes & \ablateyes & 83.33 & \textbf{87.5} & 82.35 & \textbf{80.0} & 775.29 & 33.07 \\
\bottomrule
\end{tabular*}

\end{table*}

Following the evolution procedure used in the main experiments, we vary the diagnostic information available to the Patch Agent to examine its effect on localized repair (Table~\ref{tab:diagnostic-ablation}). Summary-only receives candidate reports; Trace-only adds raw trace tools; Trace + Memory Inspection additionally exposes the store; \ourframework{} adds Patch History. No Memory and Original Provider are reference controls.

The Trace + Memory Inspection variant achieves the highest GAIA diagnostic accuracy at 86.67\%, versus 70.00\% for Trace-only and 63.33\% for Summary-only. The Patch History variant reaches 83.33\%, versus 86.67\% for Trace + Memory Inspection; the one-task difference comprises three tasks answered correctly only without history and two answered correctly only with history. Relative to Summary-only, Trace-only uses 24.4\% fewer average task steps and 7.2\% less task time. These resource differences describe downstream task execution.

\paragraph{How Patch History informs subsequent repair.}
The evolution records show how history guides later edits. In the xBench DeepSeek run, a patch raised the retrieval threshold from 0.7 to 0.8 to suppress irrelevant memories. A later repair consulted recovery and regression outcomes, recognized over-filtering, and adjusted the threshold to 0.75 while retaining protection against stale guidance. In ALFWorld, the Patch Agent recalled that lowering the threshold to 0.2 had admitted weakly related memories; it kept the current 0.3 threshold and changed the no-match fallback to return empty guidance rather than an unrelated old cheatsheet. Both subsequent revisions were installed.

Across 14 archived provider transitions, the median code-change ratio is 4.4\% and the median transition touches one function (Appendix Table~\ref{tab:patch-locality}). Appendix~\ref{app:evolution-cost} discusses the diagnosis, code-modification, and validation costs of evolution.

\begin{table}[!htbp]
\centering
\caption{A recorded GAIA repair case under different diagnostic interfaces.}
\label{tab:gaia-diagnostic-case}
\small
\setlength{\tabcolsep}{4pt}
\renewcommand{\arraystretch}{1.1}
\begin{tabular*}{\linewidth}{@{\extracolsep{\fill}}>{\raggedright\arraybackslash}p{0.20\linewidth}>{\raggedright\arraybackslash}p{0.36\linewidth}>{\raggedright\arraybackslash}p{0.38\linewidth}@{}}
\toprule
\textbf{Aspect} & \textbf{Summary-only} & \textbf{Trace-only} \\
\midrule
Evidence & Grouped failure reports & Zero-length traces; empty guidance steps \\
Proposed diagnosis & Insufficient cross-verification & Early-return guard under \texttt{MemoryStatus.IN} may suppress guidance \\
Edit & Add a verification rule to the storage prompt & Remove the early-return guard \\
\bottomrule
\end{tabular*}
\end{table}

\noindent\textbf{Evidence-grounded repair decisions.} Table~\ref{tab:gaia-diagnostic-case} contrasts two recorded GAIA repair decisions. The trace-based diagnosis identifies a concrete provider-code target, whereas the summary-based diagnosis motivates a generic storage instruction. A complementary xBench case illustrates a decision to leave the provider unchanged: the proposer records \texttt{no\_patch} after inspecting retrieved memory and noting that failed tasks had already received correct guidance. Its manifest also cites weak outcomes from a previous threshold adjustment. Together, these cases illustrate how execution evidence can inform both the location of a proposed edit and the decision not to modify the provider.

\begin{figure*}[!t]
  \centering
  \includegraphics[width=\textwidth]{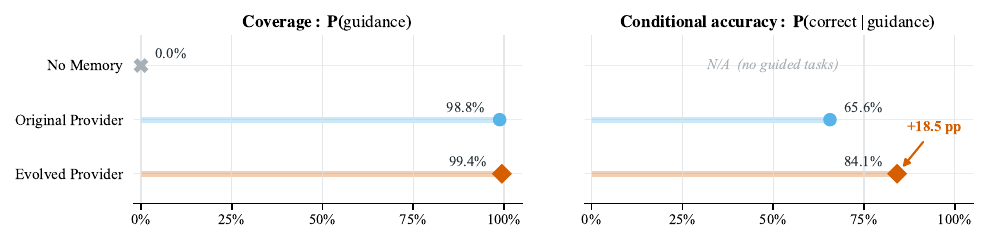}
  \caption{GAIA guidance during online memory updates. Original versus evolved provider coverage: 98.8\% versus 99.4\%; accuracy among guided tasks: 65.6\% versus 84.1\%. Code and memory state both vary.}
  \label{fig:memory-guidance-utility}
\end{figure*}

\paragraph{Guidance coverage and conditional accuracy.}
\label{sec:guidance-utility}
Figure~\ref{fig:memory-guidance-utility} separates guidance frequency from outcomes when guidance is supplied. On GAIA, coverage is 98.8\% for the original provider and 99.4\% for the evolved provider, whereas accuracy among guided tasks is 65.6\% for the original versus 84.1\% for the evolved provider. The larger observed difference is in guided-task outcomes, rather than guidance frequency; both provider code and memory state vary across the compared online trajectories.
\FloatBarrier

\FloatBarrier
\section{Conclusion}
\label{sec:conclusion}

MemScope turns evolution of an existing memory provider into trace- and memory-grounded code repair. With DeepSeek V3.2, MemScope reaches 83.64\% on GAIA and 78\% on xBench, versus 67.88\% and 70\% for MemEvolve + Flash-Searcher. Matched online comparisons favor the evolved over original DynamicCheatsheet on xBench and ALFWorld unseen tasks (71.64\% versus 50.75\% on ALFWorld). In the GAIA diagnostic ablation, trace-plus-memory accuracy is 86.67\% versus 63.33\% for summary-only. Together, these results show that localized repair can improve an existing provider during continued use.

\label{sec:main-end}
\FloatBarrier
\clearpage
\subsection*{AI use statement}

Generative AI was used to assist with experiment planning and protocol design, implementation and debugging of experimental tooling, organization and interpretation of experimental results, and drafting and revision of this manuscript, including conversion of an earlier draft to the ICLR format. The research method itself uses language models as task executors, a code-generating Patch Agent, and, for web tasks, a correctness judge, as described in the method and experimental sections. Automated checks were used to recompute the additional result tables from retained records and to check the manuscript's compilation and references. These checks do not replace author verification of claims, citations, code, or AI-assisted content. The authors are responsible for the final submitted content and for completing the corresponding disclosure in the submission form.

\bibliography{references,review_evidence/verified_related}
\bibliographystyle{iclr2027_conference}
\clearpage
\appendix
\section{Patch-Agent Repair Prompt}
\label{app:prompts}

The following is the recorded Patch-Agent provider-repair prompt template; run-specific inputs appear as placeholders.

\lstdefinestyle{promptstyle}{
  basicstyle=\scriptsize\ttfamily,
  numbers=none,
  breaklines=true,
  columns=fullflexible,
  keepspaces=true,
  showstringspaces=false,
  frame=single,
  xleftmargin=0.5em,
  xrightmargin=0.5em
}

\subsection{Patch-Agent Provider Repair}
\label{app:prompts-patch-agent}

\begin{lstlisting}[style=promptstyle,caption={Revised Patch-Agent provider-repair prompt template.},label={lst:prompt-patch-agent}]
You are MemScope PatchProposer, a coding agent for segment-local memory provider repair.

Your job is to make one minimal, evidence-backed patch to the CURRENT provider.
You are NOT designing a new provider, NOT creating a new skill, and NOT exploring a new architecture.

The current execution segment's accuracy is below the best prior-episode aggregate accuracy. Inspect the available evidence to determine whether a localized code repair is warranted.

You do NOT run the full benchmark. You may run only local syntax/contract checks through the allowed tools.
The outer trainer records subsequent task outcomes in Patch History to inform future repair.

## Inputs

Current provider identity:
- provider_name: <PROVIDER_NAME>
- provider_class_name: <PROVIDER_CLASS_NAME>
- memory_type: <MEMORY_TYPE>
- skill_name: <SKILL_NAME>
- failure_count: <FAILURE_COUNT>
- success_count: <SUCCESS_COUNT>
- evidence_mode: trace

Score context:
<SCORE_CONTEXT_JSON>

Files:
- original_provider_file: <ORIGINAL_PROVIDER_FILE>
- candidate_provider_file: <CANDIDATE_PROVIDER_FILE>
- skill_file: <SKILL_FILE>

Trace workspace index:
<TRACE_TASK_INDEX_JSON>

Use task_id values from this workspace index when calling trace tools. Do not invent task ids. If you need more detail, call list_traces or read the specific task_id from the index.

Patch history for this provider:
<PATCH_HISTORY>

Use this history as feedback. If a previous patch direction had weak follow-up outcomes, such as low fixed rate, guided_wrong, or success_regressed, do not repeat that direction unless the current evidence clearly contradicts the history. Choose a different repair direction or explicitly reject the patch.

Important:
- Edit ONLY candidate_provider_file.
- Do NOT edit original_provider_file.
- Do NOT create a new provider file.
- Do NOT create a new skill file.
- Do NOT add a v2 name, v2 class, or v2 skill.
- If a skill update is useful, write it only as learned_fix_note in patch_manifest.json.

## Available Tools

Trace tools:
- list_traces with arguments {"outcome": "all|failed|success"}
- read_trace_header with arguments {"task_id": "..."}
- read_trace_steps with arguments {"task_id": "...", "step_indices": [0, 1]}
- grep_traces with arguments {"pattern": "...", "max_matches": 10, "outcome": "all|failed|success"}

Memory inspection:
- memory_store with arguments {"action": "stats|query|list|read", "query": "...", "task_id": "...", "step_index": 0, "context": "...", "status": "BEGIN|IN", "entry_id": "...", "outcome": "all|failed|success", "limit": 10, "include_filtered": true}

Provider patch tools:
- read_candidate_provider with arguments {"max_chars": 24000}
- read_skill_file with arguments {"max_chars": 12000}
- edit_candidate_provider with arguments {"old_text": "...", "new_text": "...", "expected_occurrences": 1}
- run_candidate_checks with arguments {}
- write_patch_manifest with arguments {"manifest": {...}}

Use the bound native tools directly. Base each next action on the actual tool results returned to you. Independent reads may be requested together, but do not request edits, checks, or manifest creation before observing their prerequisites. Do not print JSON that merely describes a tool call.

If write_patch_manifest reports missing evidence, perform the named missing inspection and call write_patch_manifest again. The first rejected manifest is not a durable completion artifact. The durable completion artifact is patch_manifest.json.

## Workflow Guidance

Step 1: Inspect evidence.
- Call list_traces, then inspect at least one failed trace in detail.
- Read at least one successful trace, when available, and state what the patch must preserve.
- Prefer read_trace_steps for the smallest relevant step range.
- Ground the patch in observed trace evidence. Do not invent trace contents.
- Before proposing retrieval parameter changes, use memory_store to verify whether relevant memories exist and whether they are filtered or malformed.
- Compare raw_store_matches, selection_evidence, and provider_execution on at least one failed case and one successful case before changing shared retrieval behavior.
- task_outcome is authoritative for answer correctness. A trace status such as "success" may only mean execution completed; never treat it as a correct answer when task_outcome is "failed" or judgement is incorrect.

Step 2: Inspect current provider.
- Read candidate_provider_file before editing.
- Identify the smallest code location that can plausibly fix the observed failure pattern.

Allowed local patch scope:
- retrieval logic inside provide_memory
- ranking or filtering logic
- thresholds, top_k, score normalization, deduplication
- storage/extraction of fields already compatible with current records
- formatting of memory guidance
- defensive error handling
- small helper functions used by the existing provider

Forbidden scope:
- new provider class
- changed provider class name
- changed MemoryType
- new provider identity
- new skill identity
- new schema that invalidates existing memories
- replacing most of the provider
- unrelated refactor
- broad architecture rewrite
- dataset-specific hardcoded answers, task IDs, or benchmark names

Step 3: Patch candidate provider.
- Use edit_candidate_provider. Do not output a full replacement provider as final text.
- Prefer one small exact replacement. If multiple edits are necessary, keep them targeted.
- Preserve initialize(), provide_memory(), and take_in_memory contracts.
- take_in_memory must return tuple[bool, str].
- initialize must return bool.

Step 4: Validate locally.
- Run candidate checks after editing.
- If checks fail, fix the candidate file and re-run checks.

Step 5: Write patch_manifest.json.
- Write the manifest when you have a local patch, an explicit no-patch decision, or a new-provider handoff.
- If the required fix is architecture-level, do not edit code; write needs_new_provider=true.
- If current evidence does not justify a safe change, do not edit code; write patch_type=no_patch.

Manifest schema when patched:
{
  "patch_type": "local_patch",
  "needs_new_provider": false,
  "provider_name": "<PROVIDER_NAME>",
  "provider_class_name": "<PROVIDER_CLASS_NAME>",
  "memory_type": "<MEMORY_TYPE>",
  "skill_name": "<SKILL_NAME>",
  "evidence_mode": "trace",
  "candidate_provider_file": "<CANDIDATE_PROVIDER_FILE>",
  "edited_files": ["<CANDIDATE_PROVIDER_FILE>"],
  "trace_tools_used": ["list_traces", "..."],
  "trace_files_consulted": ["..."],
  "failed_pattern": "observed recurring behavior",
  "success_preservation": "successful behavior to preserve",
  "memory_store_diagnostic": "result of store inspection",
  "patch_history_decision": "not_repeating_failed_direction | repeating_with_reason | no_history",
  "root_cause": "one concrete code-level failure cause",
  "patch_summary": "one sentence describing the exact patch",
  "expected_effect": "why this should improve the failed segment",
  "risk": "what could regress",
  "validation": {
    "syntax_check": "passed|failed|not_run",
    "import_check": "passed|failed|not_run",
    "isolated_retrieval_check": "passed|failed|not_run"
  },
  "learned_fix_note": "short note appended to the original skill only if accepted"
}

If evidence does not justify a safe patch, use patch_type="no_patch", leave edited_files empty, and provide rejection_reason. If a local patch is insufficient, set needs_new_provider=true, leave edited_files empty, and provide new_provider_reason.

Start now.
\end{lstlisting}

\clearpage
\section{Repair Cases and Evidence}
\label{app:compact-details}

The recorded cases illustrate evidence-guided repair decisions; the table summarizes edit locality.

\subsection{GAIA Failure-Localization Case}

The GAIA case contrasts a broad storage instruction with a specific retrieval-code change. Summary-only groups two zero-step failures with three incorrect-answer cases, diagnoses insufficient cross-verification, and adds a generic verification rule to the storage prompt. Trace-only inspects \texttt{trajectory\_length}=0 and empty \texttt{memory\_guidance\_steps}, identifies an early return under \texttt{MemoryStatus.IN} as the proposed cause of missing guidance, and removes that guard. The case makes the diagnostic difference concrete: raw execution evidence identifies a code path that the summary-based proposal leaves unchanged.

\subsection{Recorded No-Patch Decision}

Table~\ref{tab:no-patch-case} gives a complementary xBench case: the Kimi proposer retains the current provider after inspecting retrieved memory, downstream failures, and a previous unsuccessful threshold change. The quoted manifest shows how the evidence can support leaving retrieval unchanged.

\begin{table}[!htbp]
\centering
\small
\caption{A recorded \texttt{no\_patch} decision in xBench.}
\label{tab:no-patch-case}
\renewcommand{\arraystretch}{1.15}
\begin{tabular}{@{}>{\raggedright\arraybackslash}p{0.32\linewidth}@{\hspace{0.03\linewidth}}>{\raggedright\arraybackslash}p{0.65\linewidth}@{}}
\toprule
\rowcolor[gray]{0.94}
\textbf{Original field} & \textbf{Recorded value} \\
\midrule
\texttt{patch\_type} & \texttt{"no\_patch"} \\
\texttt{needs\_new\_provider} & \texttt{false} \\
\texttt{edited\_files} & \texttt{[]} \\
\texttt{patch\_history\_decision} & \texttt{"not\_repeating\_failed\_direction"} \\
\bottomrule
\end{tabular}
\par\medskip
{\setlength{\fboxsep}{7pt}%
\colorbox[gray]{0.96}{%
\begin{minipage}{\dimexpr\linewidth-2\fboxsep\relax}
\textbf{Original \texttt{rejection\_reason}}\par\smallskip
No evidence-backed local provider change is justified. The 0.7 threshold is correctly blocking an irrelevant correct memory in task 28, and lowering it would risk guided\_wrong. Tasks 24 and 29 fail after correct guidance is delivered, so retrieval changes cannot address them. The previous threshold-based gating patch already had weak follow-up outcomes (success regressed), so repeating or tightening that direction is unsafe without contradictory evidence.
\end{minipage}}}%
\par\smallskip
\begin{minipage}{\linewidth}
\footnotesize\raggedright
Source: \path{cheatsheet_memory_skill_1788529914321_5e7245.patch_manifest.json}.
\end{minipage}
\end{table}

\subsection{Contrasting Retrieval Repairs}
\label{app:contrasting-repairs}

The DeepSeek and Kimi cases show that evidence-guided repair can adjust retrieval in either direction. DeepSeek's proposer attributes failures to correct-only retrieval at a 0.7 threshold and fallback to stale global cheatsheet text; its installed patch searches all records, lowers the threshold to 0.5, expands the candidate pool, and boosts correct records. Kimi's proposer observes 37 correct records among 54 and diagnoses weak cross-domain matches under permissive retrieval; its patch raises the threshold from 0.7 to 0.8. The contrast illustrates how the proposed repair depends on the inspected memory and failure pattern rather than applying a uniform threshold change.

\subsection{Edit Locality}

Table~\ref{tab:patch-locality} measures edit scope across 14 archived provider-snapshot transitions: one from Trace + Memory Inspection, six from Summary-only, and seven from Trace-only. Median changes involve 16.5 lines and one function, against median provider-size proxies of 369 lines and 16 functions. The median recorded per-transition normalized fraction is 4.3862\%, reported as 4.4\%. This fraction is computed across individual transitions, rather than by dividing the line-count medians.

\begin{table}[!t]
\caption{Syntactic changes across archived provider snapshots.}
\label{tab:patch-locality}

\centering
\small
\setlength{\tabcolsep}{8pt}
\renewcommand{\arraystretch}{1.15}
\begin{tabular}{lrr}
\toprule
\rowcolor[gray]{0.94}
\textbf{Metric} & \textbf{\makecell{Snapshot\\change}} & \textbf{\makecell{Provider-size\\proxy}} \\
\midrule
Lines & 16.5 & 369 \\
Functions & 1 & 16 \\
\addlinespace[2pt]
Median recorded normalized fraction & 4.4\% & 100\% \\
\bottomrule
\end{tabular}

\end{table}

\paragraph{Main-table baseline sources.}
The GAIA and xBench baselines in Table~\ref{tab:main-table-1} include Alita~\citep{qiu2025alita}, Langfun Agent~\citep{peng2023langfun}, TraseAgent~\citep{laws2024trase}, OpenAI Deep Research~\citep{openai2025deepresearch}, h2oGPTe~\citep{h2oai2024h2ogpte}, Desearch~\citep{desearchai2024desearch}, and OMNE~\citep{jiang2024longtermmemory}. The remaining agent frameworks are OWL~\citep{hu2025owl}, TapeAgents~\citep{bahdanau2024tapeagents}, AutoAgent~\citep{tang2025autoagent}, Magentic-One~\citep{fourney2024magenticone}, Cognitive Kernel-Pro~\citep{fang2025cognitivekernelpro}, OAgents~\citep{zhu2025oagents}, Agent KB~\citep{tang2025agentkb}, and Flash-Searcher~\citep{qin2025flash}.

\subsection{Evolution Cost Analysis}
\label{app:evolution-cost}

The comparison separates diagnosis, code modification, and post-modification validation. This stage-wise account explains where localized repair changes the work performed during evolution.

\paragraph{Diagnosis.}
Diagnosis is shared by complete architecture generation and localized repair. MemEvolve diagnoses architectural defects before redesign, and ALMA analyzes candidate designs and evaluation logs~\citep{zhang2025memevolvemetaevolutionagentmemory,xiong2026learningcontinuallylearn}. MemScope uses execution evidence and prior repair outcomes to identify a code region for modification.

\paragraph{Code modification.}
Complete generation emits a full provider implementation for each candidate. MemScope generates replacements for selected regions and reuses the rest of the implementation. The archived edit-size measurements in Table~\ref{tab:patch-locality} quantify this limited modification scope; they measure changed code rather than model-token expenditure.

\paragraph{Post-modification validation.}
MemEvolve evaluates candidate architectures on task batches before selecting parents for further evolution~\citep{zhang2025memevolvemetaevolutionagentmemory}. ALMA performs trial runs and debugging, then evaluates candidates through memory collection and deployment~\citep{xiong2026learningcontinuallylearn}. MemScope instead checks a local code update and continues task execution with the updated architecture and retained runtime memory. Patch History records the subsequent recovered failures and regressed successes so later proposals can correct an unsuccessful direction. Behavioral feedback is thus collected along a continuing evolution trajectory, rather than through a separate task evaluation for each of multiple complete candidates.

\end{document}